\documentclass[a4paper,conference]{IEEEtran}
\ifCLASSINFOpdf
\else
\fi

\usepackage{hyperref}
\usepackage{color}
\usepackage{mathrsfs}  
\usepackage{amssymb,amsmath,bm}
\usepackage{multicol,multicap,graphicx}
\usepackage{subcaption}
\usepackage{wrapfig}
\usepackage{picinpar}
\usepackage{cutwin}
\usepackage{algorithm}  
\usepackage{algorithmic}
\usepackage{amsfonts}
\usepackage{amsfonts,amssymb}
\usepackage{url}
\usepackage{stmaryrd}

\DeclareMathOperator*{\argmin}{arg\,min}

\begin{document}
%
\title{Deep Feature Registration for Unsupervised Domain Adaptation}



\author{\IEEEauthorblockN{Youshan Zhang}
\IEEEauthorblockA{ Computer Science and Artificial Intelligence, \\Yeshiva University, NYC, NY \\
Email: youshan.zhang@yu.edu}
\and
\IEEEauthorblockN{Brian D.\ Davison}
\IEEEauthorblockA{Computer Science and Engineering, \\ Lehigh University, Bethlehem, PA, USA\\
Email: bdd3@lehigh.edu}}

\maketitle

\begin{abstract}
While unsupervised domain adaptation has been explored to leverage the knowledge from a labeled source domain to an unlabeled target domain, existing methods focus on the distribution alignment between two domains. However, how to better align source and target features is not well addressed. In this paper, we propose a deep feature registration (DFR) model to generate registered features that maintain domain invariant features and simultaneously minimize the domain-dissimilarity of registered features and target features via histogram matching. We further employ a pseudo label refinement process, which considers both probabilistic soft selection and center-based hard selection to improve the quality of pseudo labels in the target domain. Extensive experiments on multiple UDA benchmarks demonstrate the effectiveness of our DFR model, resulting in new state-of-the-art performance.
\end{abstract}


%
\IEEEpeerreviewmaketitle

\section{Introduction}
The availability of a large amount of labeled training samples is usually the prerequisite of typical machine learning algorithms. However, a shortage of labeled data in the domain of interest is not uncommon in many real-world applications. Therefore, it is of great significance to transfer knowledge from one label-rich (source) domain to a label-scarce (target) domain. However, there is a domain gap issue between the two domains, which is caused by the varied data distributions of different domains. 

To alleviate the domain gap issue without manual annotation, unsupervised domain adaptation (UDA) is well explored to learn transferable representation from the labeled source domain and unlabeled target domain. Existing domain adaptation methods assume that the feature distributions of the source and target domains are different but share the same label space. Early traditional based methods include subspace learning~\cite{gong2012geodesic,zhang2019transductive}, and distribution alignment~\cite{jiang2017integration,wang2018visual,zhang2021deep2}. Subspace learning methods transfer the samples of both domains from the original feature space into a common latent subspace, which consists of the shared features across the two domains while preserving inherent geometric data structure via a lower rank or sparse representations~\cite{gong2012geodesic,zhang2019transductive}. The traditional distribution alignment usually reduces domain discrepancy via aligning  marginal~\cite{long2014transfer,jiang2017integration}, conditional~\cite{gong2016domain} and joint~\cite{zhang2017joint,wang2018visual} distributions. However, traditional DA methods highly depend on features that are extracted from raw images.  The performance of traditional methods is tremendously improved by using deep features (e.g., ResNet50~\cite{long2017deep,wang2018visual})~\cite{zhang2020impact,zhang2021efficient}. 

Recent deep neural networks show superiority in improving UDA performance.  Generally, these deep learning-based DA methods can be roughly categorized as discrepancy-based methods~\cite{meng2018adversarial} and adversarial learning-based methods~\cite{zhang2021adversarial1}. The former aligns the distributions of source and target domains by directly minimizing the different distance metrics between the two domains. The latter methods enforce the feature representations to be indistinguishable by a domain discriminator and the feature extractor tries to confuse the discriminator. The domain invariant features are expected to be extracted from the two domains. 

Although existing DA methods have admittedly achieved promising results, most features are encoded from the neural network without explicitly aligning the features of the two domains. Feature distribution of two domains is still difficult to align. 
Finally, when pseudo labeling is used on the target domain, noisy labels are problematic.
To address aforementioned challenges, our contributions are three-fold:
\begin{itemize}
    \item To explicitly align the source and target domain extracted features, we are the first to impose a feature registration loss to align these two features to generate registered features, which can maintain both source and target domain information.

    \item To further match source and target feature distributions, we enforce a histogram matching loss, which can reduce the domain discrepancy.

    \item To suppress noisy pseudo labels in the target domain, we develop an easy-to-hard refinement process that considers both probabilistic soft and center-based hard selection.  We then form a high-quality pseudo-labeled target domain so as to jointly optimize the network and improve model performance.
    
\end{itemize}

\section{Related work}
Given the popularity of deep neural networks, they have also shown great success in the UDA problem~\cite{chen2019progressive,tang2020discriminative,lee2019sliced}. The discrepancy based method is one of the most popular deep network models, and it aims to minimize the discrepancy between the source and target distributions by proposing different loss functions, such as Maximum Mean Discrepancy (MMD)~\cite{tzeng2014deep}, CORrelation ALignment~\cite{sun2016deep}, Kullback-Leibler divergence~\cite{meng2018adversarial}, Wasserstein distance~\cite{bhushan2018deepjdot} and least squares~\cite{zhang2021deep}. Recent work included the modification of these different distance functions. Kang et al.~\cite{kang2019contrastive} extends MMD to the contrastive domain discrepancy loss.  Rahman et al.~\cite{rahman2020minimum} proposes a model based on the alignment of second-order statistics (covariances) as well as maximizing the mean discrepancy of the source and target data. Sliced Wasserstein discrepancy (SWD)~\cite{lee2019sliced} utilizes the geometric 1-Wasserstein as the discrepancy measure to obtain the dissimilarity probability of source and target domains.

Adversarial methods are another popular model to reduce domain discrepancy of different domains by using an adversarial objective with a domain discriminator. Inspired by GANs~\cite{goodfellow2014generative}, adversarial learning has shown its power in learning domain invariant representations. It consists of a domain discriminator and a feature extractor. The domain discriminator aims to distinguish the source domain from the target domain, while the feature extractor aims to learn domain-invariant representations to fool the domain discriminator~\cite{ganin2016domain,tzeng2017adversarial,zhang2021adversarial1}.  Domain Adversarial Neural Network (DANN)~\cite{ghifary2014domain} is a representative work. Later many adversarial learning methods are proposed.  SymNets~\cite{zhang2019domain} maximized the discrepancy between the outputs of the two classifiers. The cycle-consistent adversarial domain adaptation CyCADA~\cite{hoffman2018cycada} implemented domain adaptation at both pixel-level and feature-level by using cycle-consistent adversarial training. TADA~\cite{wang2019transferable} was also built upon the adversarial domain adaptation framework, which added adversarial alignment constraints on both transferable local regions and global images
through two local/global attention modules.

Pseudo-labeling is another technique to improve the generalizability of the model in the target domain. Pseudo-labeling typically generates pseudo labels for the target domain based on the predicted class probability~\cite{zhang2018collaborative,chen2019progressive,zhang2021adversarial2}. Therefore, some target domain label information can be included during training, and further reduce domain divergence. In deep networks, the source classifier usually generates the pseudo labels (and then uses them as if they were real labels). Saito et al.~\cite{saito2017asymmetric} proposed an asymmetric tri-training method for UDA to generate pseudo labels for target samples using two networks while the third can learn from them to obtain target discriminative representations.  
Zhang et al.~\cite{zhang2018collaborative} designed a new criterion to select pseudo-labeled target samples and developed an incremental model (iCAN), in which they select samples iteratively and retrain the network using the expanded training set.  PFAN~\cite{chen2019progressive} aligns the discriminative features across domains progressively and develops iterative learning to generate pseudo labels. Zhang et al.~\cite{zhang2020label} offers a label propagation with augmented anchors (A2LP) method to improve the label propagation via generation of unlabeled virtual samples with high confidence label prediction. These methods highly rely on pseudo labels to compensate for the lack of categorical information in the target domain. However, they do not check the quality of pseudo-labels, as noisy pseudo-labeled samples hurt model performance. Our work differs from these approaches by generating high confidence examples using both probabilistic soft and center-based hard selection in Sec.\ref{Sec: PLR}.

\section{Methodology}
\vspace{-0.1cm}
\subsection{Problem}
In this work, we consider the unsupervised  domain adaptation (UDA) classification problem in the following setting. There exists a labeled source domain $\mathcal{D_S} = \{x_s^i, y_s^i \}_{i=1}^{n_s}$ with $n_s$ samples in $C$ categories and an unlabeled target domain $\mathcal{D_T} = \{x_t^j\}_{j=1}^{n_t}$ with $n_t$ samples in the same $C$ categories. We assume that the data in the two domains are drawn from different distributions but share the same label space. The goal of UDA is to get a well-trained classifier so that domain discrepancy is minimized and generalization error in the target domain is reduced.

In UDA, existing methods rarely update source domain features except when learning them from the trained neural network. We instead directly replace the source domain features with the registered features that can better represent shared features between both source and target domains. We further align feature distribution via histogram matching. The domain divergence can be further reduced by the pseudo labeling refinement process of the target domain.

\begin{figure}[t]
\centering
\includegraphics[width=1.\columnwidth]{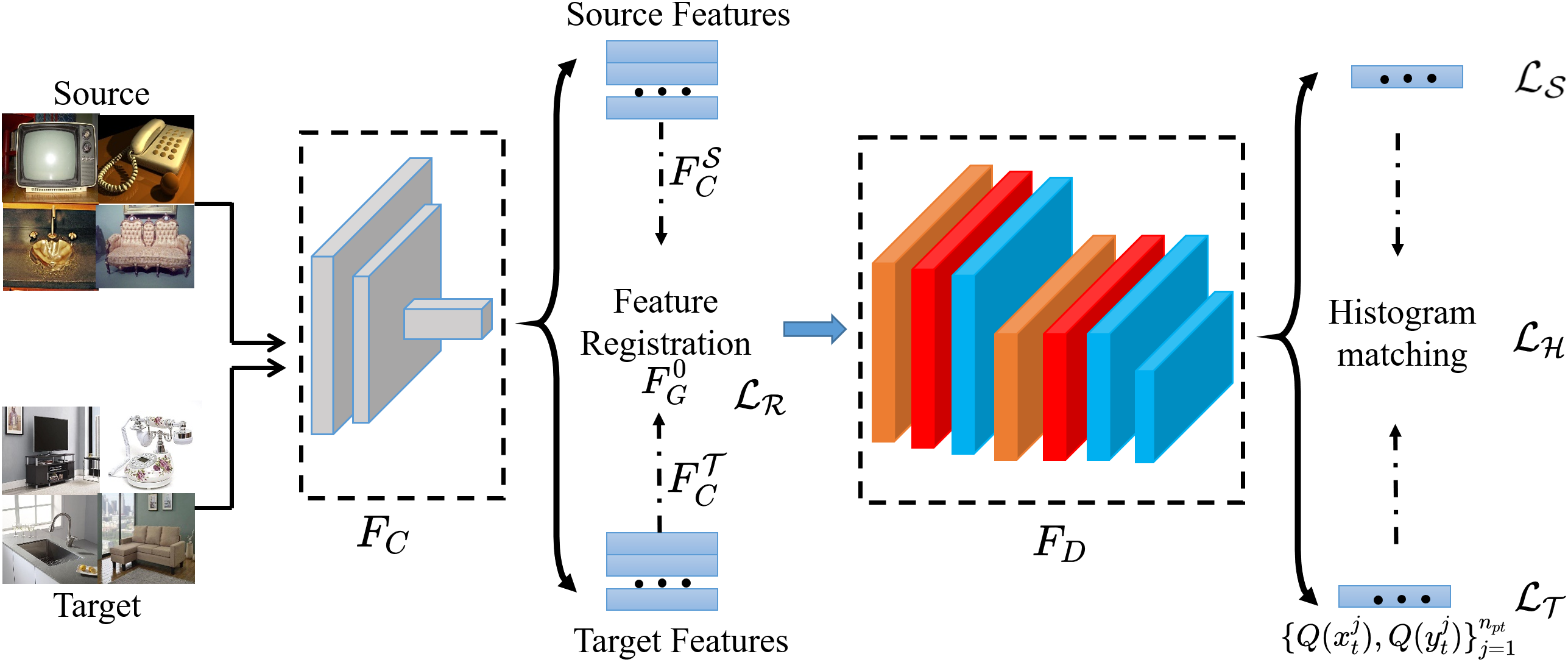}
\vspace{+0.01cm}
\caption{The architecture of the DFR model. We first employ $F_C$ to extract coarse features for two domains and then perform feature registration between source ($F_C^\mathcal{S}$) and target ($F_C^\mathcal{S}$) features. The registered features ($F_G^0$) and target features will jointly optimize shared detailed feature extractor $F_D$. $\mathcal{L_R}$ is the feature registration loss, $\mathcal{L_S}$ is the source classification loss, $\mathcal{L_H}$ is the histogram matching loss, and $\mathcal{L_T}$ is the pseudo-labeled target domain classification loss. $\{ Q(x_t^j), Q(y_t^j)\}_{j=1}^{n_{pt}}$ is the high quality pseudo-labeled target domain after $T$ times pseudo labeling refinement processes (Orange block: Linear layer, red block: ReLU layer, and cyan block: BatchNormalization layer).}  
\vspace{-0.3cm}
\label{fig:DFR}
\end{figure}

\subsection{Feature Registration}\label{Sec: FR}
We first extract coarse features for both source and target images from a shared coarse feature extractor $F_C$, which is a typical one of CNN backbones. The motivation for feature registration is to find shared invariant features (registered features $F_G^0$) between source features and target features. Given extracted batch-wise source features $B(F_C^\mathcal{S}) \in \mathbb{R}^{n_b \times d}$ and extracted batch-wise target features $B(F_C^\mathcal{T}) \in \mathbb{R}^{n_b \times d}$, where $n_b$ is the number of batch size and $d$ is extracted feature dimensionality using $F_C$, we aim to generate registered features $F_G^0$ using gradient descent. To estimate the registered features $F_G^0$, we employ a stochastic gradient descent optimizer $\mathcal{N}$ as follows,
 \begin{equation}
     F_G^0 = \mathcal{N}(B(F_C^\mathcal{S}), B(F_C^\mathcal{T}), F^0  )
 \end{equation}
where $F^0$ is the initial registered features and we initialize it as $B(F_C^\mathcal{T})- B(F_C^\mathcal{S})$ to accelerate convergence. 

To ensure the registered features $F_G^0$ contain shared target features' information, we first impose a target domain registration loss $\mathcal{L_R^T}$.      
 \begin{equation}\label{eq:lrt}
     \mathcal{L_R^T} = \sum_{p_{ij} \in \mathcal{W}} |F_G^0(p_{ij}) - B(F_C^\mathcal{T}(p_{ij}))|,
 \end{equation}
where $p_{ij}$ is the position in batch-wise feature matrix $\mathcal{W} \in \mathbb{R}^{n_b \times d} \ (i \in \{1,\cdots,n_b\}, j\in \{1,\cdots,d\}$) (for either $F_G^0$ or $B(F_C^\mathcal{T})$) and $|\cdot|$ takes the absolute value. $F_G^0$ will be close to batch-wise target features $B(F_C^\mathcal{T})$ by minimizing $\mathcal{L_R^T}$. 

In the first iteration, the $F_G^0 = F^0 = B(F_C^\mathcal{T})- B(F_C^\mathcal{S})$. The target domain registration loss in Eq.~\ref{eq:lrt} is the pixel-wise difference between registered features and the batch-wise target domain features, which is able to align the source and target domain features. However, a single $\mathcal{L_R^T}$ might lead to $F_G^0$ is very similar to  $B(F_C^\mathcal{T})$, that will lose source domain features' information. To improve the quality of $\mathcal{L_R^T}$, which contains both shared features between source and target domains, we then design source domain registration loss:
 \begin{equation}
     \mathcal{L_R^S} = \sum_{p_{ij} \in \mathcal{W}} |F_G^0(p_{ij}) - B(F_C^\mathcal{S}(p_{ij}))|.
 \end{equation}
Therefore, in feature registration, we employ the hybrid loss between the source domain registration loss and the target domain registration loss to get the registration features in the following equation.
  \begin{equation}\label{eq:lr}
  \begin{aligned}
     \mathcal{L_R}  = \mathcal{L_R^S} + \alpha \mathcal{L_R^T}  & =  \sum_{p_{ij} \in \mathcal{W}} |F_G^0(p_{ij}) - B(F_C^\mathcal{S}(p_{ij}))| + \\ & \alpha  \sum_{p_{ij} \in \mathcal{W}} |F_G^0(p_{ij}) - B(F_C^\mathcal{T}(p_{ij}))|       
  \end{aligned}
 \end{equation}
 where $\alpha$ is a balance factor between the two loss functions. $F_G^0$ is generated for each pair of source and target mini-batches. The registered features $F_G^0$ eventually maintain shared invariant features, which maps from the source domain to the target domain. 
 
\subsection{Source domain classifier}
After feature registration, registered features $F_G^0$ will be fed into the shared detailed feature extractor $F_D$, which can represent more detailed features at the categorical level. The task in the labeled source domain is formulated as:
\begin{equation}\label{eq:lc}
    \mathcal{L_S} = - \frac{1}{n_s} \sum_{\textcolor{red}{b=1}}^{n_s/n_b} \sum_{i=1}^{n_b}  \mathcal{L}_{ce} (F_D(F_{Gb}^{0i}), B_{\textcolor{red}{b}}(y_s^i)), 
\end{equation}
where $\mathcal{L}_{ce}$ is the typical cross-entropy loss. During training of shared detailed feature extractor $F_D$, the registered features $F_G^{0}$ will replace batch-wise $B(F_C^\mathcal{S})$ and keep updating via Eq.~\ref{eq:lr}, and $b$ is batch-wise data. 

\subsection{Histogram matching}
To further reduce the domain divergence between source and target domains, we employ a histogram matching step to align the distributions of the two domains at the feature level. We assume that the histogram can represent the distribution of the source and target domain features. Hence, aligning the distributions of two domains is equivalent to minimizing the difference between source and target histograms. Histogram matching usually refers to transforming the histogram of one image so that it looks like another image. One major limitation of histogram matching using original images for minimizing domain discrepancy is that it assumes that images have similar spectral properties when their pixel distributions are aligned. However, this does not hold, especially when source images are different from target images. Therefore, we perform the histogram matching at the feature level since features after the detailed features extracted by $F_D$ are already aligned to a certain extent. The basic principle of feature level histogram matching is to compute the histogram source and target features individually, then compute their discrete cumulative distribution functions (CDFs). Let $Hist(\cdot)$ be histogram generation operation, and we expect the CDF of registered features and target features are similar to each other. Therefore, we develop the following histogram matching loss:
\begin{equation}\label{eq:lh}
    \mathcal{L_H} = \sum_{h=1}^H  |Hist(F_D(F_G^{0}))_h - Hist(F_D(B(F_C^\mathcal{T})))_h|,
\end{equation}
where $H$ is the number of bins of a histogram. By minimizing histogram matching loss, we can further expect the distributions between two domains to be aligned.

\subsection{Pseudo labeling refinement}\label{Sec: PLR}
Considering no labels in the target domain, existing methods also try to generate pseudo labels for the target domain to improve the robustness of their model. However, the detrimental effects of bad pseudo-labels are still significant. To mitigate this issue, we employ a $T$ times recurrent easy-to-hard pseudo-label refinement process to improve the quality of the pseudo-labels in the target domain, which considers both probabilistic soft selection and center-based hard selection. 

The initial shared $F_D$ is optimized by $\mathcal{L_S}$ and $\mathcal{L_H}$. For the inference, we can directly get predicted results for one target domain sample $F_D (F_C(x_t^j))$. Let $\text{Softmax} (F_D (F_C(x_t^j)))$ be the predicted probability for each class, and $y_{pt}^j = max(\text{Softmax} (F_D (F_C(x_t^j))))_{index}$ be its dominant class label, where $max (\cdot)_{index}$ return the index of the maximum probability value. Therefore, for the probabilistic soft selection, a higher quality pseudo label is defined as $max(\text{Softmax} (F_D (F_C(x_t^j)))) > p_t$, where $p_t$ is a threshold probability in number of $t$ training. For $T$ times recurrent easy-to-hard pseudo-label refinement, for easy examples, $p_t$ has a higher value and for hard examples, $p_t$ has a lower value, hence $p_1 > p_2 > \cdots > p_T$.

For center-based hard selection, let $\{M_S^c \in \mathbb{R}^{1 \times C}\}_{c=1}^C$ be the $C$ source cluster centers in the feature space after $F_D$. Let $y_s==c$ mean all indices of source label class $c$; then all source class $c$ features are $F_D (F_C(x_s^{y_s==c}))$ and $n_c$ is the number of its features. We define the class center as: $M_S^c = \frac{1}{n_c} \sum_{ii=1}^{n_c} F_D (F_C(x_s^{ii}))$. For any target sample $x_t^j$, we compute its distance to each source domain cluster center $M_S^c$ as $dist_j = \{|F_D (F_C(x_t^j)) - M_S^c|_{L1}\}_{c=1}^C \in \mathbb{R}^{1 \times C}$, where $|\cdot|_{L1}$ is the L1 norm. Therefore, a higher quality pseudo label exists if $min (dist_j)_{index} == max(\text{Softmax} (F_D (F_C(x_t^j))))_{index}$, where $min (dist_j)_{index}$ returns the index of the cluster center with minimum distance. Hence, the center-based hard selection ensures the target sample has the same prediction target label as the closest source domain center from the shared $F_D$. 

In pseudo labeling refinement, we combine probabilistic soft and center-based hard selection together to form a robust new high quality pseudo-labeled domain $\{ Q(x_t^j), Q(y_t^j)\}_{j=1}^{n_{pt}}$ in the following equation,
\begin{equation}\label{eq:plr}
\begin{aligned}
     & \{ Q(x_t^j), Q(y_t^j)\}_{j=1}^{n_{pt}} \quad \text{if and  only if} \\ & (\text{Softmax} (F_D (F_C(x_t^j))) > p_t)  \ \&  (min (dist_j)_{index}\\ & == max(\text{Softmax} (F_D (F_C(x_t^j))))_{index}),
\end{aligned}
\end{equation}
where $Q(x_t)$ and $Q(y_t)$ represent the high quality target domain samples and pseudo labels, $n_{pt}$ is the number of higher quality pseudo labels for the target domain. We hence can mitigate detrimental effects of bad pseudo-labels using Eq.~\ref{eq:plr}. Similar to Eq.~\ref{eq:lc}, we define the pseudo-labeled target domain loss as: 
\begin{equation}\label{eq:lt}
    \mathcal{L_T} = - \frac{1}{n_{pt}} \sum_{b=1}^{n_{pt}/n_b} \sum_{j=1}^{n_b}  \mathcal{L}_{ce} (F_D(B_b(F_C(Q(x_t^j)))), B_b(Q(y_t^j))),
\end{equation}
where $B (\cdot)$ is also the batch-wise data.

\subsection{DFR model}
Fig.~\ref{fig:DFR} depicts the overall framework of our proposed DFR model. Considering Sec.~\ref{Sec: FR} to Sec.~\ref{Sec: PLR}, our model minimizes the following objective function:
\begin{equation}\label{eq:loss_all}
  \mathop{\argmin}  \   (\mathcal{L_R} + \mathcal{L_S}   + \beta \mathcal{L_H} + \sum_{t=1}^{T} \mathcal{L}_{\mathcal{T}}^{t}  )    
\end{equation}
where $\mathcal{L_R}$ is the feature registration loss, $\mathcal{L_S}$ is the source classification loss, $\mathcal{L_H}$ is the histogram matching loss, and $\mathcal{L}_{\mathcal{T}}$ is the pseudo-labeled target domain classification loss, and we repeat the pseudo labeling refinement process $T$ times. $\beta$ controls the weight of histogram matching loss.  The overall training algorithm of our DFR model is shown in  Alg.~\ref{alg:DFR}.

\begin{algorithm}[h]
   \caption{Deep Feature Registration Network. $B(\cdot)$ denotes the mini-batch training sets, $I$ is the number of iterations. $T$ is the number of refinement steps.}
   \label{alg:DFR}
\begin{algorithmic}[1]
   \STATE {\bfseries Input:} labeled source samples  $\mathcal{D_S} = \{x_s^i, y_s^i \}_{i=1}^{n_s}$ and unlabeled target samples $\mathcal{D_T} = \{x_t^j\}_{j=1}^{n_t}$
   \STATE {\bfseries Output:} predicted target domain labels 
   \REPEAT
   \STATE Derive $B(F_C^\mathcal{T})$ and $ B(F_C^\mathcal{S})$ sampled from $\mathcal{D_S}$ and $\mathcal{D_T}$ after finely tuned $F_C$
   \FOR{$iter =1$ {\bfseries to} $I$}
   \FOR{$t=1$ {\bfseries to} $T$}
   \STATE Generate registered features $F_G^0$ using Eq.~\eqref{eq:lr} 
   \STATE Optimize $F_D$ using Eq.~\eqref{eq:lc} and Eq.~\eqref{eq:lh}
   \STATE Form  pseudo-labeled target domain $\{ Q(x_t^j), Q(y_t^j)\}_{j=1}^{n_{pt}}$ using Eq.~\eqref{eq:plr}
   \STATE Refine $F_D$ using Eq.~\eqref{eq:lt}
   \ENDFOR
   \ENDFOR
   \UNTIL{converged}
\STATE Make prediction for target domain samples based on trained $F_C$ and $F_D$.
\end{algorithmic}
\end{algorithm}

\section{Experiments}
\subsection{Experimental Setup}
 \paragraph{Datasets.}We evaluate our model on three popular benchmark image datasets:  Office-31, Office-Home, and VisDA-2017. \textbf{Office-31} \cite{saenko2010adapting} has 4,110 images from three  domains: Amazon (A), Webcam (W), and DSLR (D) in 31 classes. In experiments, A$\shortrightarrow$W represents transferring knowledge from domain A to domain W.  \textbf{Office-Home}~\cite{venkateswara2017deep} dataset contains 15,588 images from four domains: Art (Ar), Clipart (Cl), Product (Pr), and Real-World (Rw) in 65 classes. \textbf{VisDA-2017}~\cite{peng2017visda} is a particularly challenging dataset due to a large domain-shift between the synthetic images (152,397 images from VisDA) and the real images (55,388 images from COCO) in 12 classes. We test our model on the setting of synthetic-to-real as the source-to-target domain.

 \paragraph{Implementation details.}
We implement our approach using PyTorch and extract features for the three datasets from finely tuned ($F_C$) ResNet50 (Office-31, Office-Home) and ResNet101 (VisDA-2017) networks~\cite{he2016deep}. The 1,000 features are then extracted from the last fully connected layer for the source and target features. The outputs of three Linear layers are 512, 256 and $|C|$, respectively, where $C$ is the number of classes in each dataset. Parameters in recurrent pseudo labeling are  $T=3$ and $\{p_t\}_{t=1}^{3} = [0.9, 0.6, 0.3]$. Optimizer (Adam), learning rate ($0.001$), batch size (64), $\alpha =0.6$, $\beta =0.01$ and number of epochs (210) are determined by performance on the source domain. In histogram matching, 10 bins are used. More detailed parameter analysis is presented in supplementary material.  We compare our results with 20 state-of-the-art methods. For a fair comparison, we directly report results from original papers. Experiments are performed with an Nvidia GeForce 1080 Ti.

\begin{table*}[h!]
\begin{center}
\small
\captionsetup{font=small}
\caption{Accuracy (\%) on Office-Home dataset (based on ResNet50)}
\vspace{-0.1cm}
\setlength{\tabcolsep}{+0.1mm}{
\begin{tabular}{rccccccccccccc}
\hline \label{tab:OH}
Task & Ar$\shortrightarrow$Cl &  Ar$\shortrightarrow$Pr & Ar$\shortrightarrow$Rw & Cl$\shortrightarrow$Ar & Cl$\shortrightarrow$Pr & Cl$\shortrightarrow$Rw & Pr$\shortrightarrow$Ar & Pr$\shortrightarrow$Cl & Pr$\shortrightarrow$Rw & Rw$\shortrightarrow$Ar & Rw$\shortrightarrow$Cl & Rw$\shortrightarrow$Pr & \textbf{Ave.}\\
\hline
TAT~\cite{liu2019transferable} &51.6  &69.5 & 75.4 & 59.4 & 69.5 & 68.6 & 59.5 &50.5 &76.8 &70.9 &56.6 &81.6 &65.8 \\
TADA~\cite{wang2019transferable} & 53.1 &72.3& 77.2& 59.1 &71.2& 72.1& 59.7& 53.1& 78.4 &72.4& \textbf{60.0} &82.9& 67.6 \\
SymNets~\cite{zhang2019domain} & 47.7 & 72.9 & 78.5 & 64.2  & 71.3  &74.2  & 64.2  & 48.8 &  79.5&  74.5 &52.6 & 82.7& 67.6 \\
DMP~\cite{luo2020unsupervised} & 52.3 &73.0 &77.3 &64.3 &72.0 &71.8 &63.6 &52.7 &78.5 &72.0 &57.7 &81.6 & 68.1 \\
DCAN~\cite{li2020domain} & 54.5 &75.7 &81.2 &67.4 &74.0 &76.3 &67.4 &52.7 &80.6 &74.1 &59.1 &83.5 &70.5\\ 
SRDC~\cite{tang2020unsupervised} &52.3 &76.3 &81.0 &69.5 &76.2 &78.0 &68.7 &53.8 &81.7 &76.3 &57.1 &85.0 & 71.3 \\
ESD~\cite{zhang2021enhanced} &53.2 &75.9 &82.0 &68.4 &79.3 &79.4 &69.2 &54.8 &81.9 &74.6 &56.2 &83.8 &71.6 \\
PICSCS~\cite{li2023pseudo} &56.0 &79.0 &81.0 &67.6 &81.3 &79.9 &68.4 &55.0 &\textbf{82.4} &72.3 &58.5 &85.0 & 72.2\\
\hline
\hline
\textbf{DFR}& \textbf{56.2} &	\textbf{81.2}  &	\textbf{83.6}  &	\textbf{70.4} &	\textbf{82.2}  & \textbf{80.3} & \textbf{71.1} &	\textbf{56.3} &	82.3 &	\textbf{76.5} &	58.4  &	\textbf{86.2} & 	\textbf{73.7} \\
\hline
\end{tabular}}
\vspace{-0.3cm}
\end{center}
\end{table*}

\begin{table*}[h!]
\begin{center}
\small
\captionsetup{font=small}
\caption{Accuracy (\%) on VisDA-2017 dataset (based on ResNet101)}
\vspace{-0.1cm}
\setlength{\tabcolsep}{+1.1mm}{
\begin{tabular}{rccccccccccccc}
\hline \label{tab:VisDA}
Task & plane& bcycl& bus& car& horse& knife& mcycl& person& plant& sktbrd& train& truck & \textbf{Ave.}\\
\hline
Source-only~\cite{he2016deep} &  55.1 &53.3 &61.9& 59.1& 80.6& 17.9& 79.7& 31.2& 81.0& 26.5& 73.5& 8.5 & 52.4 \\
DANN~\cite{ghifary2014domain} 	&81.9 &77.7& 82.8 &44.3& 81.2& 29.5 &65.1 &28.6 & 51.9 & 54.6 & 82.8 & 7.8 & 57.4\\
DAN~\cite{long2015learning}	& 87.1& 63.0& 76.5& 42.0 &90.3& 42.9 &85.9 &53.1& 49.7 &36.3& 85.8 &20.7 &61.1\\
JAN~\cite{long2017deep}	& 75.7& 18.7& 82.3 &86.3& 70.2 &56.9& 80.5& 53.8 &92.5 &32.2& 84.5& 54.5 &65.7 \\
DMP~\cite{luo2020unsupervised} &92.1 &75.0 &78.9 &75.5 &91.2 &81.9 &89.0 &77.2 &93.3 &77.4 &84.8 &35.1 &79.3 \\
DADA~\cite{tang2020discriminative} & 92.9 &74.2& 82.5& 65.0& 90.9& 93.8& 87.2& 74.2& 89.9& 71.5& 86.5 &48.7&  79.8 \\
STAR~\cite{lu2020stochastic} & 95.0& 84.0& 84.6& 73.0& 91.6 &91.8& 85.9 &78.4& 94.4& 84.7 &87.0 &42.2& 82.7 \\
CAN~\cite{kang2019contrastive} & \textbf{97.9} & 87.2& 82.5& 74.3 & 97.8 & 96.2& 90.8& 80.7& 96.6 &96.3 &87.5& 59.9& 87.2 \\
CDCL~\cite{luo2023adversarial} & 97.3 &\textbf{90.5} &83.2 &59.9 &96.4 &\textbf{98.4} &91.5 &\textbf{85.6} &96.0 &95.8 &\textbf{92.0} &\textbf{63.8 }&87.5\\
\hline
\hline
\textbf{DFR}&   97.2 & 89.6 & \textbf{85.1} & \textbf{77.6} & \textbf{98.5} & 96.7 & \textbf{91.6} & 82.2 & \textbf{96.8} & \textbf{96.8} & 88.4 & 60.3 & \textbf{88.4} \\
\hline
\end{tabular}}
\vspace{-0.3cm}
\end{center}
\end{table*}

\begin{table}[!htb]
\small
\captionsetup{font=small}
      \caption{Accuracy (\%) on Office-31 dataset (based on ResNet50)}
      \vspace{-0.1cm}
      \centering
\setlength{\tabcolsep}{+0.3mm}{
\begin{tabular}{rcccccccc|c|c|c|c|c|c|c|c|}
\hline \label{tab:O31}
Task & A$\shortrightarrow$W &  A$\shortrightarrow$D & W$\shortrightarrow$A & W$\shortrightarrow$D & D$\shortrightarrow$A & D$\shortrightarrow$W  & \textbf{Ave.}\\
\hline
GSM~\cite{zhang2019transductive}	& 84.8  & 82.7  & 73.5   &96.6  & 70.9  & 95.0  & 83.9\\
JAN~\cite{long2017deep}&	85.4 &	84.7	&70.0 &	99.8	&68.6 &	97.4 &	84.3\\
DMP~\cite{luo2020unsupervised} & 93.0 & 91.0 & 70.2 & \textbf{100} & 71.4 & 99.0 & 87.4\\
TADA~\cite{wang2019transferable} &94.3 & 91.6  & 73.0  &99.8  & 72.9 & 98.7 & 88.4 \\
SymNets~\cite{zhang2019domain}& 90.8& 93.9  &72.5  & \textbf{100} & 74.6 & 98.8& 88.4 \\
TAT~\cite{liu2019transferable} & 92.5 & 93.2 & 73.1 & \textbf{100}& 73.1 & 99.3 & 88.5 \\
CAN~\cite{kang2019contrastive} & 94.5 & 95.0  &77.0   &99.8  & 78.0 & 99.1 &  90.6 \\
SRDC~\cite{tang2020unsupervised} & 95.7  & 95.8  &77.1  & \textbf{100} &76.7  & 99.2& 90.8\\ 
PICSCS~\cite{li2023pseudo} & 93.2& 93.6  &78.0 &\textbf{100} &77.1& 99.1 &90.2\\
\hline
\hline
\textbf{DFR} & \textbf{95.9} &	\textbf{96.3} &	\textbf{80.6} &	\textbf{100} &	\textbf{79.6} &	99.3 &	\textbf{92.0}	\\
\hline
\end{tabular}}
\vspace{-0.2cm}
\end{table}

\begin{table}[!htb]
\small
\captionsetup{font=small}
      \caption{Ablation experiments on Office-31 dataset}
      \vspace{-0.1cm}
      \centering
\setlength{\tabcolsep}{+0.3mm}{
\begin{tabular}{lcccccccc}
\hline \label{tab:ab}
Task & A$\shortrightarrow$W &  A$\shortrightarrow$D & W$\shortrightarrow$A & W$\shortrightarrow$D & D$\shortrightarrow$A & D$\shortrightarrow$W  & \textbf{Ave.}\\
\hline
DFR$-$H/T/R	& 85.7 &	89.4 &	75.0 &	97.1 &	75.8 &	95.1 &	86.4\\
DFR$-$R/T  & 86.9 &	90.1 &	75.3&	97.6 &	76.9 &	96.1&	87.2\\
DFR$-$H/R &91.2 &	90.9 &	77.2  &	98.0 &	77.8 &	96.3 &	88.6\\	
DFR$-$H/T 	& 91.5 & 91.8 &	77.6 &	98.3 &	78.3 &	96.4 &	89.0\\ 
DFR$-$R	& 93.5 & 94.1&	78.2&	99.0 &	78.5 &	97.0&	90.1\\
DFR$-$T & 94.8 &	95.6 &	79.0 &	99.0  &	79.0 &	97.2&	90.8\\
DFR$-$H & 95.5 &	96.0 &	79.4 &	99.2 &	79.2 &	98.4 &	91.3	\\
\hline
\hline
\textbf{DFR} & \textbf{95.9} &	\textbf{96.3} &	\textbf{80.6} &	\textbf{100} &	\textbf{79.6} &	\textbf{99.3} &	\textbf{92.0}	\\
\hline
\end{tabular}}
\vspace{-0.6cm}
\end{table}

\subsection{Results}\label{sec: re}
The results on Office-Home,  VisDA-2017, and Office-31 datasets are shown in Tables~\ref{tab:OH}-\ref{tab:O31}. Overall, our DFR model outperforms all methods in terms of average accuracy and in almost all tasks of the three datasets. Notably, our DFR model substantially improves classification accuracy on difficult adaptation tasks (e.g., W$\shortrightarrow$A, D$\shortrightarrow$A task in the Office-31 dataset and the challenging VisDA-2017 and Office-Home datasets, which have a larger number of categories and domain discrepancy). For the Office-31 dataset, the mean accuracy is 92.0\%, which achieves state-of-the-art results. Compared with the best baseline (SRDC), our DFR model provides a 1.2\%  improvement. Notably, DFR wins all tasks except for W$\shortrightarrow$A task. Similarly, in Office-Home and VisDA-2017 datasets, our model is almost ahead of all tasks except Pr$\shortrightarrow$Cl task and plane category. For the Office-Home dataset, DFR achieves the highest average, 73.7\%, which is 1.5\% higher than the PICSCS method and 2.4\% higher than the DCAN model. For the VisDA-2017 dataset, the DFR model has a 0.9\% improvement over the best baseline (CDCL). Therefore, our proposed feature registration and histogram matching are useful, and the easy-to-hard refinement process is effective in improving classification accuracy.

\begin{figure*}[h]
\centering
\captionsetup{font=small}
\includegraphics[width=2.1\columnwidth]{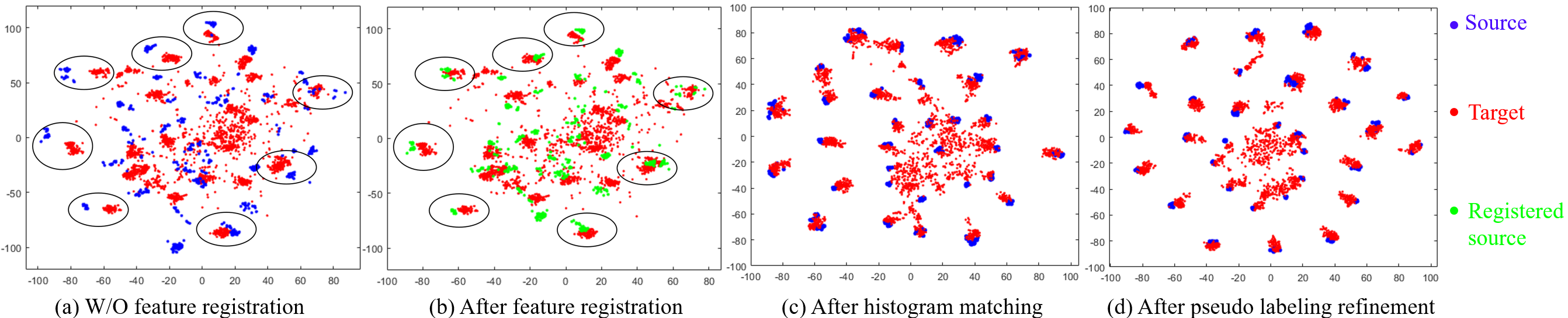}
\vspace{-0.3cm}
\caption{Feature visualization using a 2D t-SNE view of task D$\shortrightarrow$A in Office-31 dataset. Without (a) and with (b) feature registration (before $F_D$). We highlight obvious changes using black ellipses. The registered features have less dispersivity after performing feature registration. (c): after performing histogram matching and (d): after performing pseudo labeling refinement step.  (blue color: source features, red color: target features, green color: registered features based on source domain). Best viewed in color.} 
\label{fig:tsne2}
\end{figure*}

\begin{figure*}[t]
\centering
\captionsetup{font=small}
\includegraphics[width=2\columnwidth]{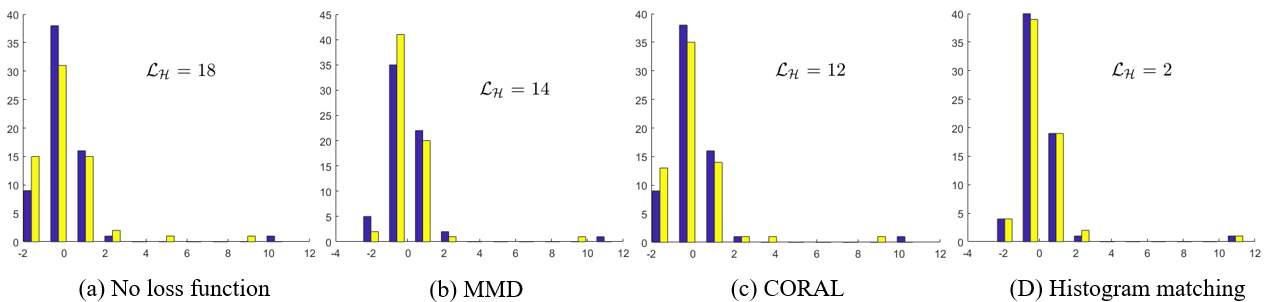}
\caption{Histogram comparison of one sample in task Pr$\shortrightarrow$Rw in Office-Home dataset.  Our histogram matching strategy achieves the smallest difference between source and target feature histograms. (blue color: source domain feature histogram, yellow color: target feature histogram). }
\label{fig:hist}
\end{figure*}

\subsection{Ablation study}\label{sec:ab}
To demonstrate the effects of different loss functions on final classification accuracy, we conduct an ablation study w.r.t. to each component of our proposed DFR model in Tab.~\ref{tab:ab}.  For this, we reuse ResNet50 on the Office-31 dataset of twelve tasks. R represents feature registration loss $\mathcal{L_R}$, H is the histogram matching loss $\mathcal{L_H}$,  and T is the pseudo-labeled target domain loss  $\mathcal{L_T}$. Notice that source classification loss $\mathcal{L_S}$ is required for UDA. ``DFR$-$H/T/R” is implemented without $\mathcal{L_H}$, $\mathcal{L_T}$, and $\mathcal{L}_\mathcal{R}$. It is a simple model, which only reduces the source risk without minimizing the domain discrepancy using $\mathcal{L_S}$. ``DFR$-$H/T” only performs the feature registration process. ``DFR$-$H” reports results without performing histogram matching. We can find that with the increasing number of loss functions, the accuracy of our model keeps improving. The effective of loss functions on classification accuracy is ordered as $\mathcal{L}_\mathcal{R} >\mathcal{L_{T}} > \mathcal{L_H}$. Therefore, our feature registration, histogram matching, and easy-to-hard target domain pseudo labeling refinement approaches are effective in minimizing target domain risk and improving the accuracy.

\subsection{Discussion}
From results in Sec.~\ref{sec: re}, our DFR always achieves the highest average accuracy, outperforming SOTA methods. There are three compelling reasons. First, we perform feature registration, which can map the source features to the target features and minimize the domain discrepancy before $F_D$. To better understand this, we show the 2D t-SNE feature visualization without and with feature registration in Fig.~\ref{fig:tsne2} using the task of D$\shortrightarrow$A in the Office-31 dataset. Fig.~\ref{fig:tsne2}(a) is the extracted source and target features from $F_C$ without performing feature registration. After minimizing $\mathcal{L_R}$, we get all registered features based on source domain in Fig.~\ref{fig:tsne2}(b), while target features are not changed (red color). Clearly, the points in the highlighted black ellipses in Fig.~\ref{fig:tsne2}(b) are closer to each other than that in Fig.~\ref{fig:tsne2}(a). Secondly, we propose a histogram matching method to further align the distribution between two domains. To show the advantages of our proposed histogram matching loss $\mathcal{L_H}$, we compare the histogram of another two well-known distance-based methods (MMD~\cite{long2015learning} and CORAL~\cite{sun2016deep} in Fig.~\ref{fig:hist}. We directly report the value of $\mathcal{L_H}$, which can represent the difference between source and target histograms. A smaller $\mathcal{L_H}$ means a smaller domain discrepancy between two domains. We find that our histogram matching strategy has the smallest value among all other methods.  Thirdly, we develop an easy-to-hard refinement process to improve the quality of pseudo labels in the target domain. This strategy considers both probabilistic soft and center-based hard selection, and it hence can push the shared $F_D$ towards the target domain. As discussed in Sec.~\ref{sec:ab}, the $T$ times easy-to-hard refinement process is effective in improving the classification accuracy and further reduces the domain discrepancy. 
We also show t-SNE feature visualization after histogram matching and pseudo labeling refinement in Fig.~\ref{fig:tsne2}(c) and Fig.~\ref{fig:tsne2}(d). We can find that Fig.~\ref{fig:tsne2}(d) is our DFR model, and it can learn highly discriminative features and keep clear boundaries, especially in the central portion.

\vspace{-0.3cm}
\section{Conclusion}
We have proposed a novel deep feature registration (DFR) model for UDA. To align source and target domain features, we develop a feature registration loss to minimize the intrinsic discrepancy between two domain features. To further align domain distributions, we design a histogram matching loss. We also employ an easy-to-hard refinement process, which combines both probabilistic soft and center-based hard selection and can improve the quality of pseudo labels.
Extensive experiments demonstrate that the proposed DFR model achieves higher accuracy than state-of-the-art methods.






%



\bibliographystyle{unsrt}
\bibliography{references}

\begin{thebibliography}{10}

\bibitem{gong2012geodesic}
B.~Gong, Y.~Shi, F.~Sha, and K.~Grauman.
\newblock Geodesic flow kernel for unsupervised domain adaptation.
\newblock In {\em Proceedings of IEEE Conference on Computer Vision and Pattern
  Recognition (CVPR)}, pages 2066--2073. IEEE, 2012.

\bibitem{zhang2019transductive}
Y.~Zhang, S.~Xie, and B.~D. Davison.
\newblock Transductive learning via improved geodesic sampling.
\newblock In {\em Proceedings of the 30th British Machine Vision Conference},
  2019.

\bibitem{jiang2017integration}
M.~Jiang, W.~Huang, Z.~Huang, and G.~G. Yen.
\newblock Integration of global and local metrics for domain adaptation
  learning via dimensionality reduction.
\newblock {\em IEEE Transactions on Cybernetics}, 47(1):38--51, 2017.

\bibitem{wang2018visual}
J.~Wang, W.~Feng, Y.~Chen, H.~Yu, M.~Huang, and P.~S. Yu.
\newblock Visual domain adaptation with manifold embedded distribution
  alignment.
\newblock In {\em Proceedings of the 26th ACM International Conference on
  Multimedia}, MM '18, pages 402--410, 2018.

\bibitem{zhang2021deep2}
Y.~Zhang and B.~D. Davison.
\newblock Deep spherical manifold gaussian kernel for unsupervised domain
  adaptation.
\newblock In {\em Proceedings of the IEEE/CVF Conference on Computer Vision and
  Pattern Recognition Workshop}, pages 4443--4452, 2021.

\bibitem{long2014transfer}
M.~Long, J.~Wang, G.~Ding, J.~Sun, and P.~S. Yu.
\newblock Transfer joint matching for unsupervised domain adaptation.
\newblock In {\em Proceedings of the IEEE Conference on Computer Vision and
  Pattern Recognition}, pages 1410--1417, 2014.

\bibitem{gong2016domain}
M.~Gong, K.~Zhang, T.~Liu, D.~Tao, C.~Glymour, and B.~Sch{\"o}lkopf.
\newblock Domain adaptation with conditional transferable components.
\newblock In {\em Proceedings of the International Conference on Machine
  Learning}, pages 2839--2848, 2016.

\bibitem{zhang2017joint}
J.~Zhang, W.~Li, and P.~Ogunbona.
\newblock Joint geometrical and statistical alignment for visual domain
  adaptation.
\newblock In {\em Proceedings of the IEEE Conference on Computer Vision and
  Pattern Recognition}, pages 1859--1867, 2017.

\bibitem{long2017deep}
M.~Long, H.~Zhu, J.~Wang, and M.~I. Jordan.
\newblock Deep transfer learning with joint adaptation networks.
\newblock In {\em Proceedings of the 34th International Conference on Machine
  Learning}, volume~70, pages 2208--2217. JMLR.org, 2017.

\bibitem{zhang2020impact}
Y.~Zhang and B.~D. Davison.
\newblock Impact of {ImageNet} model selection on domain adaptation.
\newblock In {\em Proceedings of the IEEE Winter Conference on Applications of
  Computer Vision Workshops}, pages 173--182, 2020.

\bibitem{zhang2021efficient}
Y.~Zhang and B.~D. Davison.
\newblock Efficient pre-trained features and recurrent pseudo-labeling in
  unsupervised domain adaptation.
\newblock In {\em Proceedings of the IEEE/CVF Conference on Computer Vision and
  Pattern Recognition Workshop}, pages 2719--2728, 2021.

\bibitem{meng2018adversarial}
Z.~Meng, J.~Li, Y.~Gong, and B.~Juang.
\newblock Adversarial teacher-student learning for unsupervised domain
  adaptation.
\newblock In {\em 2018 IEEE International Conference on Acoustics, Speech and
  Signal Processing (ICASSP)}, pages 5949--5953. IEEE, 2018.

\bibitem{zhang2021adversarial1}
Y.~Zhang, H.~Ye, and B.~D. Davison.
\newblock Adversarial reinforcement learning for unsupervised domain
  adaptation.
\newblock In {\em Proceedings of the IEEE/CVF Winter Conference on Applications
  of Computer Vision}, pages 635--644, 2021.

\bibitem{chen2019progressive}
C.~Chen, W.~Xie, W.~Huang, Y.~Rong, X.~Ding, Y.~Huang, T.~Xu, and J.~Huang.
\newblock Progressive feature alignment for unsupervised domain adaptation.
\newblock In {\em Proc.\ IEEE Conf.\ on Computer Vision and Pattern
  Recognition}, pages 627--636, 2019.

\bibitem{tang2020discriminative}
H.~Tang and K.~Jia.
\newblock Discriminative adversarial domain adaptation.
\newblock In {\em Proceedings of the AAAI Conference on Artificial
  Intelligence}, volume~34, pages 5940--5947, 2020.

\bibitem{lee2019sliced}
C.~Lee, T.~Batra, M.~H. Baig, and D.~Ulbricht.
\newblock Sliced wasserstein discrepancy for unsupervised domain adaptation.
\newblock In {\em Proceedings of the IEEE Conference on Computer Vision and
  Pattern Recognition}, pages 10285--10295, 2019.

\bibitem{tzeng2014deep}
E.~Tzeng, J.~Hoffman, N.~Zhang, K.~Saenko, and T.~Darrell.
\newblock Deep domain confusion: Maximizing for domain invariance.
\newblock {\em arXiv preprint arXiv:1412.3474}, 2014.

\bibitem{sun2016deep}
B.~Sun and K.~Saenko.
\newblock Deep coral: Correlation alignment for deep domain adaptation.
\newblock In {\em Proc.\ of European Conference on Computer Vision}, pages
  443--450. Springer, 2016.

\bibitem{bhushan2018deepjdot}
B.~Bhushan~Damodaran, B.~Kellenberger, R.~Flamary, D.~Tuia, and N.~Courty.
\newblock Deepjdot: Deep joint distribution optimal transport for unsupervised
  domain adaptation.
\newblock In {\em Proceedings of the European Conference on Computer Vision
  (ECCV)}, pages 447--463, 2018.

\bibitem{zhang2021deep}
Y.~Zhang and B.~D. Davison.
\newblock Deep least squares alignment for unsupervised domain adaptation.
\newblock {\em Proceedings of the 32th British Machine Vision Conference},
  2021.

\bibitem{kang2019contrastive}
G.~Kang, L.~Jiang, Y.~Yang, and A.~G. Hauptmann.
\newblock Contrastive adaptation network for unsupervised domain adaptation.
\newblock In {\em Proc.\ of the IEEE Conference on Computer Vision and Pattern
  Recognition}, pages 4893--4902, 2019.

\bibitem{rahman2020minimum}
M.~M. Rahman, C.~Fookes, M.~Baktashmotlagh, and S.~Sridharan.
\newblock On minimum discrepancy estimation for deep domain adaptation.
\newblock In {\em Domain Adaptation for Visual Understanding}, pages 81--94.
  Springer, 2020.

\bibitem{goodfellow2014generative}
I.~Goodfellow, J.~Pouget-Abadie, M.~Mirza, B.~Xu, D.~Warde-Farley, S.~Ozair,
  A.~Courville, and Y.~Bengio.
\newblock Generative adversarial nets.
\newblock In {\em Advances in Neural Information Processing Systems}, pages
  2672--2680, 2014.

\bibitem{ganin2016domain}
Y.~Ganin, E.~Ustinova, H.~Ajakan, P.~Germain, H.~Larochelle, F.~Laviolette,
  M.~Marchand, and V.~Lempitsky.
\newblock Domain-adversarial training of neural networks.
\newblock {\em The Journal of Machine Learning Research}, 17(1):2096--2030,
  2016.

\bibitem{tzeng2017adversarial}
E.~Tzeng, J.~Hoffman, K.~Saenko, and T.~Darrell.
\newblock Adversarial discriminative domain adaptation.
\newblock In {\em Proceedings of the IEEE Conference on Computer Vision and
  Pattern Recognition}, pages 7167--7176, 2017.

\bibitem{ghifary2014domain}
M.~Ghifary, W.~B. Kleijn, and M.~Zhang.
\newblock Domain adaptive neural networks for object recognition.
\newblock In {\em Proceedings of the Pacific Rim International Conference on
  Artificial Intelligence}, pages 898--904. Springer, 2014.

\bibitem{zhang2019domain}
Y.~Zhang, H.~Tang, K.~Jia, and M.~Tan.
\newblock Domain-symmetric networks for adversarial domain adaptation.
\newblock In {\em Proceedings of the IEEE Conference on Computer Vision and
  Pattern Recognition}, pages 5031--5040, 2019.

\bibitem{hoffman2018cycada}
J.~Hoffman, E.~Tzeng, T.~Park, J.~Zhu, P.~Isola, K.~Saenko, A.~Efros, and
  T.~Darrell.
\newblock Cycada: Cycle-consistent adversarial domain adaptation.
\newblock In {\em International conference on machine learning}, pages
  1989--1998. PMLR, 2018.

\bibitem{wang2019transferable}
X.~Wang, L.~Li, W.~Ye, M.~Long, and J.~Wang.
\newblock Transferable attention for domain adaptation.
\newblock In {\em Proceedings of the AAAI Conference on Artificial
  Intelligence}, volume~33, pages 5345--5352, 2019.

\bibitem{zhang2018collaborative}
W.~Zhang, W.~Ouyang, W.~Li, and D.~Xu.
\newblock Collaborative and adversarial network for unsupervised domain
  adaptation.
\newblock In {\em Proceedings of the IEEE Conference on Computer Vision and
  Pattern Recognition}, pages 3801--3809, 2018.

\bibitem{zhang2021adversarial2}
Y.~Zhang and B.~D. Davison.
\newblock Adversarial continuous learning in unsupervised domain adaptation.
\newblock In {\em Pattern Recognition. ICPR International Workshops and
  Challenges: Virtual Event, January 10--15, 2021, Proceedings, Part II}, pages
  672--687. Springer International Publishing, 2021.

\bibitem{saito2017asymmetric}
K.~Saito, Y.~Ushiku, and T.~Harada.
\newblock Asymmetric tri-training for unsupervised domain adaptation.
\newblock {\em arXiv preprint arXiv:1702.08400}, 2017.

\bibitem{zhang2020label}
Y.~Zhang, B.~Deng, K.~Jia, and L.~Zhang.
\newblock Label propagation with augmented anchors: A simple semi-supervised
  learning baseline for unsupervised domain adaptation.
\newblock In {\em European Conference on Computer Vision}, pages 781--797.
  Springer, 2020.

\bibitem{saenko2010adapting}
K.~Saenko, B.~Kulis, M.~Fritz, and T.~Darrell.
\newblock Adapting visual category models to new domains.
\newblock In {\em Proceedings of the European Conference on Computer Vision},
  pages 213--226. Springer, 2010.

\bibitem{venkateswara2017deep}
H.~Venkateswara, J.~Eusebio, S.~Chakraborty, and S.~Panchanathan.
\newblock Deep hashing network for unsupervised domain adaptation.
\newblock In {\em Proceedings of the IEEE Conference on Computer Vision and
  Pattern Recognition}, pages 5018--5027, 2017.

\bibitem{peng2017visda}
X.~Peng, B.~Usman, N.~Kaushik, J.~Hoffman, D.~Wang, and K.~Saenko.
\newblock Visda: The visual domain adaptation challenge.
\newblock {\em arXiv preprint arXiv:1710.06924}, 2017.

\bibitem{he2016deep}
K.~He, X.~Zhang, S.~Ren, and J.~Sun.
\newblock Deep residual learning for image recognition.
\newblock In {\em Proceedings of the IEEE Conference on Computer Vision and
  Pattern Recognition (CVPR)}, pages 770--778, 2016.

\bibitem{liu2019transferable}
H.~Liu, M.~Long, J.~Wang, and M.~Jordan.
\newblock Transferable adversarial training: A general approach to adapting
  deep classifiers.
\newblock In {\em International Conference on Machine Learning}, pages
  4013--4022, 2019.

\bibitem{luo2020unsupervised}
Y.~Luo, C.~Ren, D.~Dao-Qing, and H.~Yan.
\newblock Unsupervised domain adaptation via discriminative manifold
  propagation.
\newblock {\em IEEE Transactions on Pattern Analysis and Machine Intelligence},
  2020.

\bibitem{li2020domain}
S.~Li, C.~H. Liu, Q.~Lin, B.~Xie, Z.~Ding, G.~Huang, and J.~Tang.
\newblock Domain conditioned adaptation network.
\newblock In {\em AAAI}, pages 11386--11393, 2020.

\bibitem{tang2020unsupervised}
Hui Tang, Ke~Chen, and Kui Jia.
\newblock Unsupervised domain adaptation via structurally regularized deep
  clustering.
\newblock In {\em Proceedings of the IEEE/CVF conference on computer vision and
  pattern recognition}, pages 8725--8735, 2020.

\bibitem{zhang2021enhanced}
Y.~Zhang and B.~D. Davison.
\newblock Enhanced separable disentanglement for unsupervised domain
  adaptation.
\newblock In {\em 2021 IEEE International Conference on Image Processing
  (ICIP)}, pages 784--788. IEEE, 2021.

\bibitem{li2023pseudo}
Lei Li, Jun Yang, Yulin Ma, and Xuefeng Kong.
\newblock Pseudo-labeling integrating centers and samples with consistent
  selection mechanism for unsupervised domain adaptation.
\newblock {\em Information Sciences}, 628:50--69, 2023.

\bibitem{long2015learning}
M.~Long, Y.~Cao, J.~Wang, and M.~I. Jordan.
\newblock Learning transferable features with deep adaptation networks.
\newblock {\em arXiv preprint arXiv:1502.02791}, 2015.

\bibitem{lu2020stochastic}
Z.~Lu, Y.~Yang, X.~Zhu, C.~Liu, Y.~Song, and T.~Xiang.
\newblock Stochastic classifiers for unsupervised domain adaptation.
\newblock In {\em Proceedings of the IEEE/CVF Conference on Computer Vision and
  Pattern Recognition}, pages 9111--9120, 2020.

\bibitem{luo2023adversarial}
Xin Luo, Wei Chen, Zhengfa Liang, Chen Li, and Yusong Tan.
\newblock Adversarial style discrepancy minimization for unsupervised domain
  adaptation.
\newblock {\em Neural Networks}, 157:216--225, 2023.

\end{thebibliography}

\end{document}